# Predicting consumer-technology ownership without a diffusion history

Irina Vartanova[1,2,3], Niels Selling[3], Jennifer Viberg Johansson[4], Pontus Strimling[1,3]

[1]Institute for Futures Studies, Stockholm, Sweden
[2]Department of Women's and Children's Health, Uppsala University, Uppsala, Sweden
[3]Institute for Analytical Sociology, Linköping University, Norrköping, Sweden
[4]Centre for Research Ethics and Bioethics, Uppsala University, Uppsala, Sweden
Corresponding author: irina.vartanova@iffs.se

## Abstract

We test whether the perceived attributes of a consumer technology predict how widely it is owned. In a 2022 Prolific survey of US adults (n = 678), respondents rated 65 consumer technologies on six attributes. We then elicited the same ratings from two frontier language models, Anthropic Claude Opus 4.7 and OpenAI GPT-5.5. We regress ownership prevalence on four UTAUT2 acceptance attributes plus a log-age covariate with a sign-constrained penalized regression and evaluate it by holding out one technology at a time. The attribute model improves on a baseline of years-since-launch: mean absolute error falls by 17% with the human ratings, and by more with either model, most with Opus 4.7. Over the short 2022-to-2025 window, where ownership moved little, the same attributes do not improve on a no-change baseline. We set out the limitations of the approach, including the possibility that language-model ratings reflect prior knowledge of these technologies rather than independent attribute reasoning. We include a deployment illustration: 2027 ownership predictions for eleven products launched in 2025 and 2026.

## Introduction

The most useful adoption forecasts are available before a product reaches the market, because the investment and regulatory decisions that depend on them are made before the first reliable sales record exists. The current generation of LLM-driven consumer technologies has sharpened that pre-launch window: products that emerged in the past two years would have appeared, on most lists from three years ago, as research demonstrations rather than household goods. New-product forecasting is a long-standing problem in the diffusion literature, and its most demanding form remains unsolved, namely prediction for a focal technology with no observed adoption history of its own. We ask how the attributes of a consumer technology, the kind that can be evaluated before launch, relate to how widely it is owned. We fit an attribute-based mapping across many technologies and validate it on held-out ones.

A diffusion curve carries two kinds of information that need not be predicted together: the level of adoption, and its movement over time. For a product with no history of its own, neither can be estimated in the usual way, because there is no trajectory to fit, and even the partial early data from a just-launched product do not pin the curve's parameters down. The present study does not reconstruct a trajectory. Instead, it relates the fixed, pre-launch properties of a technology to an observed level of adoption, estimated across many technologies at once rather than inferred from the history of any single product. This addresses the first part of the problem, not the whole of it. It concerns the level at which a technology is adopted, not how that level moves over time. The movement is the part that current data and the identifiability of the curve do not yet let us reach.

The present study asks two questions. First, can a compact attribute battery drawn from the Unified Theory of Acceptance and Use of Technology 2 (UTAUT2), requiring no historical ownership trace of the focal technology, predict the population ownership prevalence of a held-out consumer technology? Second, does the answer depend on whether the rater is a survey of human respondents or a frontier language model? We collected ownership and attribute ratings on 65 consumer technologies in a 2022 Prolific survey of US adults, with a 2025 follow-up survey that re-elicited ownership. We elicited the six attributes from three sources: the 2022 human raters, Anthropic Claude Opus 4.7, and OpenAI GPT-5.5. A sign-constrained penalized regression on the four-attribute main battery (performance expectancy, hedonic motivation, effort expectancy, and social influence), evaluated by leave-one-technology-out prediction, predicts held-out prevalence at a mean absolute error of 7.1 pp on Opus 4.7, 7.5 pp on GPT-5.5, and 12.0 pp on the human raters, against 14.4 pp for a covariate-only baseline of log technology age. The same battery does not improve on a no-change baseline for the 2022-to-2025 difference. The paper closes with a deployment illustration, in which the main model is projected onto eleven products launched in 2025 and 2026 to generate 2027 ownership predictions that a later study could evaluate against actual ownership.

## Background

The dominant tradition in the technology-adoption literature explains observed diffusion rather than predicting a curve that does not yet exist. The Bass model and its descendants (Bass, 1969; Bass, Krishnan, & Jain, 1994; Mahajan, Muller, & Bass, 1990) estimate an S-shaped adoption curve from a sales series long enough to identify the peak of yearly adoption. The curve's parameters are not pinned down by partial early data (Heeler & Hustad, 1980; Srinivasan & Mason, 1986). Methods that project future diffusion from the Bass family therefore depend on existing trajectories that resemble the focal product, and the pre-launch case is exactly the one in which such trajectories are not available.

Existing approaches to pre-launch forecasting fall into analogical and subjective families (Goodwin et al., 2014; Lee et al., 2014; Mahajan et al., 1990). The analogical approach, also called "guessing by analogy," identifies a previously launched product similar to the focal product and assumes that the new product will follow a comparable S-shaped trajectory (Green & Armstrong, 2007; Kim et al., 2013; Martino, 1993; Wright & Stern, 2015). Recent extensions add information that compensates for limitations of the analogue itself, including pre-release search traffic for the focal product (Schaer, Kourentzes, & Fildes, 2026) and conditional generative architectures that combine static product descriptors with retrieved reference trajectories from historical products (Avogaro et al., 2024; Zhou et al., 2026). Across these variants, the analogical approach faces recurring limitations: unclear criteria for selecting analogues and selection bias from missing data on unsuccessful products (Goodwin et al., 2013; Grimm et al., 2018; Jiang et al., 2006). A more basic limit is the absence of any comparable product, most acute precisely when a forecast is most valuable, that is, when the focal innovation is genuinely new and no credible analogue pool exists.

Subjective approaches instead rely on managers and other experts forecasting from their expertise, or ask potential customers directly about their purchase intentions (Bosetti et al., 2012; Mahajan et al., 1990; Van Ittersum & Feinberg, 2010). Recent work has introduced more elaborate variants on both sides, including immersive virtual-reality store environments designed to elicit revealed rather than stated consumer preferences (Harz, Hohenberg, & Homburg, 2022) and structured Bayesian expert elicitation tied to specific technologies (Grimm et al., 2018). The underlying weaknesses persist. Expert judgement is susceptible to bias and overconfidence (Feiler & Tong, 2022). Consumer intention surveys overestimate real-world adoption, reflecting well-documented discrepancies between stated intentions and actual behavior (Goodwin et al., 2014; Morwitz, 2014). Every method in this family, whether expert panel, intent measure, conjoint, or virtual-reality elicitation, requires a per-product fielded study, which does not scale when many candidate technologies must be screened in parallel.

A third strand relates an adoption outcome to product attributes across a panel of technologies, then projects the fitted mapping onto a new product. Lee, Kim, Park, and Kang (2014) estimated Bass parameters for a set of products and regressed those parameters on expert-rated product and market attributes. Yamamura and colleagues (2022) addressed the same problem by a different route, combining expert domain knowledge with machine learning to forecast the market share of incremental new products in the automobile industry. These two studies are the closest precedent for the present work. They differ from it in two ways. Their predictors are expert-rated product and market attributes or other expert-supplied inputs rather than perception-based ratings, and each draws its panel from a single product category. They also differ in what they predict: Lee estimates the parameters of a diffusion curve, whereas Yamamura predicts a market-share level, which is closer in kind to the ownership prevalence examined here.

UTAUT2 supplies the attribute framework we use (Venkatesh, Thong, & Xu, 2012). UTAUT2 extends UTAUT, which Venkatesh, Morris, Davis, and Davis (2003) consolidated from the Technology Acceptance Model (Davis, 1989) and seven related models. It sits within a wider diffusion tradition. Rogers's (1983) five perceived attributes of an innovation (relative advantage, complexity, compatibility, observability, trialability) and Tornatzky and Klein's (1982) meta-analytic synthesis converge on essentially the same construct space, in which performance expectancy maps onto relative advantage and effort expectancy onto the inverse of complexity. The framework defines seven perceived constructs that drive an individual's adoption of a technology: performance expectancy (perceived usefulness), effort expectancy (perceived ease of use), social influence (perception that important others endorse using it), facilitating conditions (perceived availability of resources and support), hedonic motivation (perceived fun), price value (perceived benefit relative to cost), and habit (automaticity of use).

The UTAUT2 battery has been used overwhelmingly to model attitudes toward a single technology and individual adoption decisions, and its seven-construct space has been examined in systematic reviews and large-scale meta-analyses across consumer-technology settings (Tamilmani, Rana, Wamba, & Dwivedi, 2021; Blut, Chong, Tsigna, & Venkatesh, 2022). Engström, Vartanova, Viberg Johansson, Persson, and Strimling (2024) applied the model across 26 online recommender systems. Of perceived performance expectancy, effort expectancy, and hedonic motivation, only perceived performance expectancy was associated with recommender-system usage across applications, accounting for a substantial share of the across-application variation ($R^2 = 0.30$). The Engström et al. design moves UTAUT2 from a within-technology to a between-technology register, although it is restricted to a single product category and to three of the seven UTAUT2 constructs.

Attribute-based forecasting at any useful scale requires attribute ratings for every candidate technology in the pool, whether existing or pre-launch. The standard route is a survey in which human raters score each technology on each item, but fielding a fresh survey for every candidate does not scale. Large language models are a candidate substitute, because they produce attribute scores in seconds and apply the same item wording across an arbitrary number of technologies. Whether LLM ratings can predict technology diffusion is an open empirical question, and the evidence on LLMs as raters of social-science constructs is mixed. Studies that use LLMs as synthetic survey respondents find variance compression and prompt sensitivity in the resulting estimates of population means and regression coefficients (Bisbee et al., 2024; Goli & Singh, 2024). A published evaluation finds that the best single LLM correlates with aggregate human judgement at $r \approx 0.43$ against a human split-half reliability of 0.60, that ensembles outperform individual models, and that general-purpose model rankings do not predict which model is the better simulator of human aggregates (Nguyen, Watts, & Whiting, 2026). On a different task, predicting a population benchmark of everyday-norm judgements, the strongest proprietary models exceed even the best individual human, though their errors are correlated across runs and across models, so hybrid human-and-model ensembles outperform either source alone (Eriksson, Karlsson, Vartanova, & Strimling, 2026). A concurrent consumer-simulation benchmark audits thirteen frontier generators (including Gemini-3.1-Pro, GPT-5.2, and Claude-4.6) against real consumer-reaction criteria and reports that the best model covers only 47.8% of those criteria, indicating that current frontier LLMs remain bounded substitutes for survey panels even at the top of the capability scale (Wang, Li, & Lin, 2026). These evaluations all score LLM ratings against further human judgements rather than against measured behavior. The present design scores them directly against self-reported ownership of held-out technologies, and elicits ratings from two frontier models in different families (Anthropic Claude Opus 4.7 and OpenAI GPT-5.5) rather than committing to one.

# Methods

## The 2022 survey: technologies, instrument, and respondents

We collected primary data in two separate Prolific surveys administered on Qualtrics to US adults aged 18 and older between March and April 2022. The first survey covered 25 AI-embedded consumer products, and the second covered 40 non-AI consumer products. Each survey comprised a main collection and a short follow-up collection that added late-identified technologies and increased coverage of low-awareness products. Together the two surveys cover 65 products introduced to the market over the preceding two decades. Table S1 reports the full list, sorted from most to least owned, with each product's instrument (AI-embedded or non-AI), market-introduction year, 2022 and 2025 ownership, 2022 awareness, and per-technology rating-sample size.

Each respondent first completed two checklists covering every product in their survey, one for awareness ("Please indicate if you know what they are") and one for ownership ("Please mark all products owned by you or anyone in your household"). They then rated a random subset of the products they had marked as known, scoring each on every attribute.

The attribute battery rated six UTAUT2 acceptance constructs (performance expectancy, effort expectancy, hedonic motivation, social influence, facilitating conditions, and price value; Venkatesh, Thong & Xu, 2012). We exclude habit, the seventh UTAUT2 construct, because it reflects how much a technology is used rather than whether it has been adopted. Each construct is measured by three survey items that we averaged into a single composite score, with every item rated 1 (*Strongly disagree*) to 7 (*Strongly agree*). To measure the characteristics of each technology rather than the disposition of the individual rater, we adapted the standard items so that respondents judged whether people in general would agree with each statement, rather than reporting their own experience. This also allows us to estimate the attributes of rarely owned products. Only respondents who recognize a technology rate it, so first-person items, answerable only by owners, would have left the least-owned products with too few raters. The cost is that the battery measures perceived social consensus rather than first-person experience. The measurement structure of the battery for this survey is validated by a two-level confirmatory factor analysis (Jak et al., 2013; Table S2). The main model reported in the results uses four of the six rated constructs, performance expectancy, hedonic motivation, effort expectancy, and social influence, and reports price value and facilitating conditions as supplement add-ons (Figure S1). Full item wording appears in Table S3.

Adding either price value or facilitating conditions back to the main model moves held-out mean absolute error only within the bootstrap interval (Figure S1), and each is left out for a reason specific to the forecasting goal. Price value overlaps with the other acceptance attributes, because judging whether a product is good value means weighing what it offers against what it costs, so adding it captures part of their signal and contributes little of its own. On these data its mean absolute correlation with the other acceptance attributes runs from 0.47 to 0.57 across rating sources, and leaving it out lets the model forecast before a market price is set, which a not-yet-launched technology does not yet have. Facilitating conditions is excluded on the same reliability grounds, joined by a property of the construct itself. It is the least reliably measured construct in the battery, barely distinguished from effort expectancy, with the measurement evidence reported in the same confirmatory factor analysis (Table S2). It is also partly a consequence of diffusion rather than a fixed property of a technology, because facilitating conditions, the perceived availability of support, accessories, and compatible products, accumulates around a product precisely because the product is already widely owned, which makes it the predictor least available for a genuinely new technology, one with no installed base.

The 2022 surveys yielded 678 eligible respondents (mean age 40.6 years, SD 14.8; 65% women), and respondents in the rating loop rated on average 4.3 of their known technologies. Per-technology rating samples ranged from n = 9 to n = 90 (median n = 37). The study collected no identifying or sensitive information and did not require formal ethics approval. Informed consent was obtained from every participant.

## The 2025 follow-up survey

In January 2025 we re-fielded both surveys on Prolific, covering the same 65 technologies and re-eliciting the awareness and ownership checklists only, to track three-year movement in the diffusion outcome. The 2025 surveys yielded 683 respondents (mean age 40.2 years, SD 13.4; 66% women), which is similar in age and gender composition to the 2022 survey.

## LLM-elicited attribute ratings

We obtained the same attribute battery from two frontier large language models, Anthropic Claude Opus 4.7 and OpenAI GPT-5.5, in 2026, after ownership had been collected in 2022 and 2025. We queried each model at high reasoning effort, with item wording held fixed and no web access or tool use, producing one rating per technology-by-attribute cell. By 2026 the broad commercial record of most of these technologies, that is, which became common and which did not, was public and within the models' training data. That record does not include the present outcome, because the Prolific ownership shares predicted here have never been released. The concern is therefore not that a model could recite the outcome, but that general knowledge of how a technology fared might influence its attribute ratings, leaving them dependent on the ownership they predict. The next section reports a direct check. We otherwise treat the three sources symmetrically, as three estimates of the same attributes, and return to the rater ordering in the Discussion.

## A knowledge-cutoff check on the model ratings

A model's knowledge runs only to a fixed training cutoff, so a later release may have learned how a recently launched product fared while an earlier release could not. If the ratings encode that knowledge, prior knowledge of outcomes makes a directional prediction: relative to the earlier release, the later one should rate a known success higher on the attributes associated with success, namely usefulness, enjoyment, and value, and a known failure lower. We elicited the full six-construct battery at each release, so the prediction can be checked on every attribute, including effort expectancy, which carries the most weight in the main model, and facilitating conditions, the supplement add-on whose availability for a new technology is most in question. We re-elicited the battery from an earlier and a later release of each family (Claude Opus 4 versus Opus 4.8; GPT-5 versus GPT-5.5), with settings held fixed so that any shift reflects added knowledge rather than added capability. The technologies were sixteen products launched in 2024 and early 2025, each given by name, brand, and a short description, alongside the 65 training technologies. Because the 65 reached the market between 2002 and 2018, both releases already know which of them became common, so the movement between the earlier and later releases is a baseline for version-to-version differences unrelated to any outcome.

## Statistical analysis

**Outcomes.** The primary outcome is the share of respondents owning each technology, with the (technology, year) cell as the unit of analysis (130 observations across 65 technologies and two timepoints). A secondary main-text analysis takes the 2022-to-2025 change in ownership as the outcome, one row per technology (n = 65).

**Model.** The model predicts ownership from the acceptance constructs, scored by a given rating source, plus one covariate: log technology age at the survey date, computed from the year consumer-grade products in the category first reached the US market. We fit a sign-constrained binomial elastic-net (mixing parameter 0.5; Friedman, Hastie & Tibshirani, 2010) on aggregated ownership counts (owners and respondents per technology) in R (R Core Team, 2025). The four acceptance attributes and the log-age covariate are constrained non-negative, the UTAUT-theoretical direction for the attributes. All predictors are standardized (z-scored) within each leave-one-out training fold, using only that fold's means and standard deviations, so that no information from the held-out technology enters its own prediction. We do not include a separate dummy for the 2025 survey. Every technology is exactly three years older in 2025 than in 2022, so log age already carries the difference between the two timepoints, and when we did add the dummy its coefficient never exceeded 0.05 logits in any cell.

**Held-out evaluation.** We measure predictive performance by leave-one-technology-out (LOO) out-of-sample prediction, which mirrors the target use of predicting a technology that was not part of the fitting sample. For each of the 65 technologies, the 2022 and 2025 cells are held out together, the model is re-estimated on the remaining 128 cells with inner 10-fold cross-validation selecting the penalty that minimizes cross-validated deviance, and the two held-out cells are predicted. The main metric is across-technology mean absolute error (MAE) on predicted-minus-observed prevalence in percentage points. We compute 95% bootstrap intervals (B = 1,000) that resample technologies, with both cells of any sampled technology entering each replicate together. We apply the same procedure to a covariate-only baseline (log technology age) and a predict-training-mean baseline. Attribute-versus-baseline comparisons use a paired bootstrap on the per-technology MAE difference. Re-selecting the penalty by the one-standard-error rule rather than the cross-validated minimum leaves the rater ordering unchanged, at 7.90 pp for Opus, 8.23 pp for GPT, and 13.16 pp for the human raters.

**Change-prediction estimator.** For the 2022-to-2025 change, the unit is a single technology and the outcome is the logit-difference of ownership share, $\text{logit}\ (d_{25} + \varepsilon) - \text{logit}\ (d_{22} + \varepsilon)$ with $\varepsilon = 0.005$. Predictors are the same four attributes (scored by the rater's single rating round), the 2022 ownership level on the logit scale as anchor, and log age at 2022. We estimate a sign-constrained Gaussian elastic-net, with the penalty selected the same way, by inner cross-validation nested within leave-one-technology-out. We convert predicted logit-differences back to predicted change in percentage points for reporting.

**Permutation null.** As a sanity check on the main cell, we randomly pair each technology with another technology's attribute vector (preserving the two-timepoint structure of the donor), re-run the nested LOO procedure, and record the resulting MAE. We repeat the procedure 100 times on the Opus 4.7 cell.

# Results

## Cross-sectional ownership prevalence

Figure 1 reports leave-one-technology-out MAE for the predict-training-mean baseline (**M0**), a covariate-only baseline of log technology age (**M1**), and three versions of the four-attribute model, one per rating source (**M1 + attributes**). The four-attribute models bring MAE down from 14.44 pp (M1) and 15.08 pp (M0) to 11.95 pp on the human raters, 7.55 pp on GPT-5.5, and 7.09 pp on Opus 4.7. The best rater, Opus 4.7, gives a 51% MAE reduction over M1.

The MAE differences are paired by technology, so we test whether the attribute model improves on the covariate-only baseline M1 with a paired bootstrap on the MAE difference within each resampled technology set. The Opus cell improves on M1 by 7.35 pp (95% CI [4.77, 10.09]), and the GPT cell improves on M1 by 6.88 pp (95% CI [4.41, 9.40]). The human-rater cell's mean paired difference against M1 is 2.48 pp (95% CI [0.51, 4.37]), with the bootstrap distribution placing 99% of its mass on the human attribute model improving over M1. The Opus-versus-human rater difference is 4.87 pp (95% CI [2.30, 7.54]). To check whether the gain over M1 reflects information in the attribute values rather than added flexibility from fitting five predictors instead of one, we randomly pair technologies with attributes on the Opus 4.7 cell and re-run the entire nested LOO 100 times. Every random pairing yielded a higher MAE than the observed 7.09 pp (mean 14.76 pp, 95% CI [14.19, 15.31]). Adding either price value or facilitating conditions back into the main battery (per rating source) moves MAE only within the bootstrap interval. The two add-on cells are reported in Supplementary Figure S1.

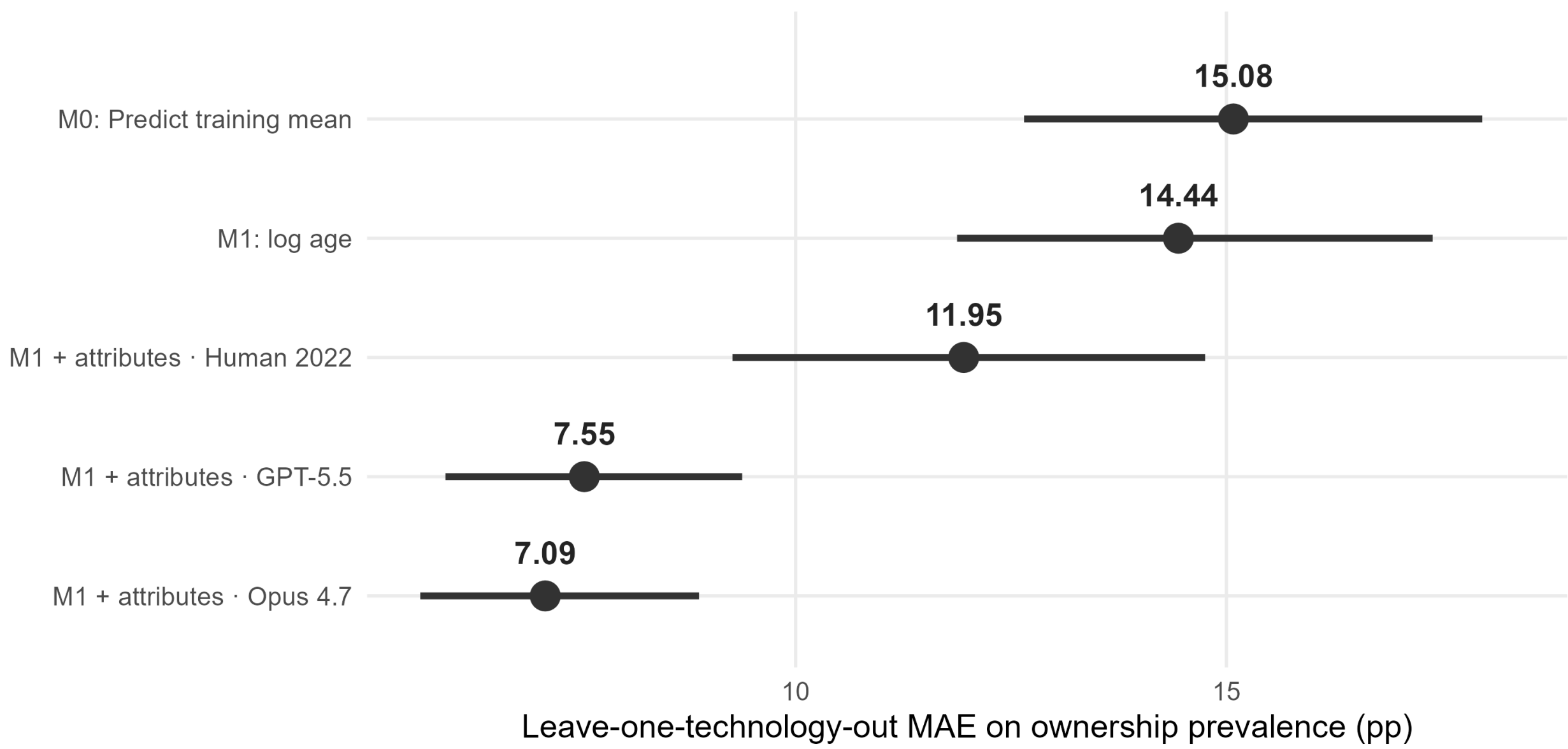

Figure 1: Leave-one-technology-out mean absolute error on Prolific 2022+2025 ownership prevalence. M0: predict the training-mean prevalence. M1: log technology age (single covariate). M1 + attributes: M1 plus the four UTAUT2 acceptance attributes scored by, respectively, the 2022 human raters, GPT-5.5, and Opus 4.7. Bars are 95% bootstrap CIs over technology resamples.

Figure 2 plots observed against predicted prevalence per (technology, year) cell under the leave-one-out scheme, one panel per rating source, with the two survey years shown separately.

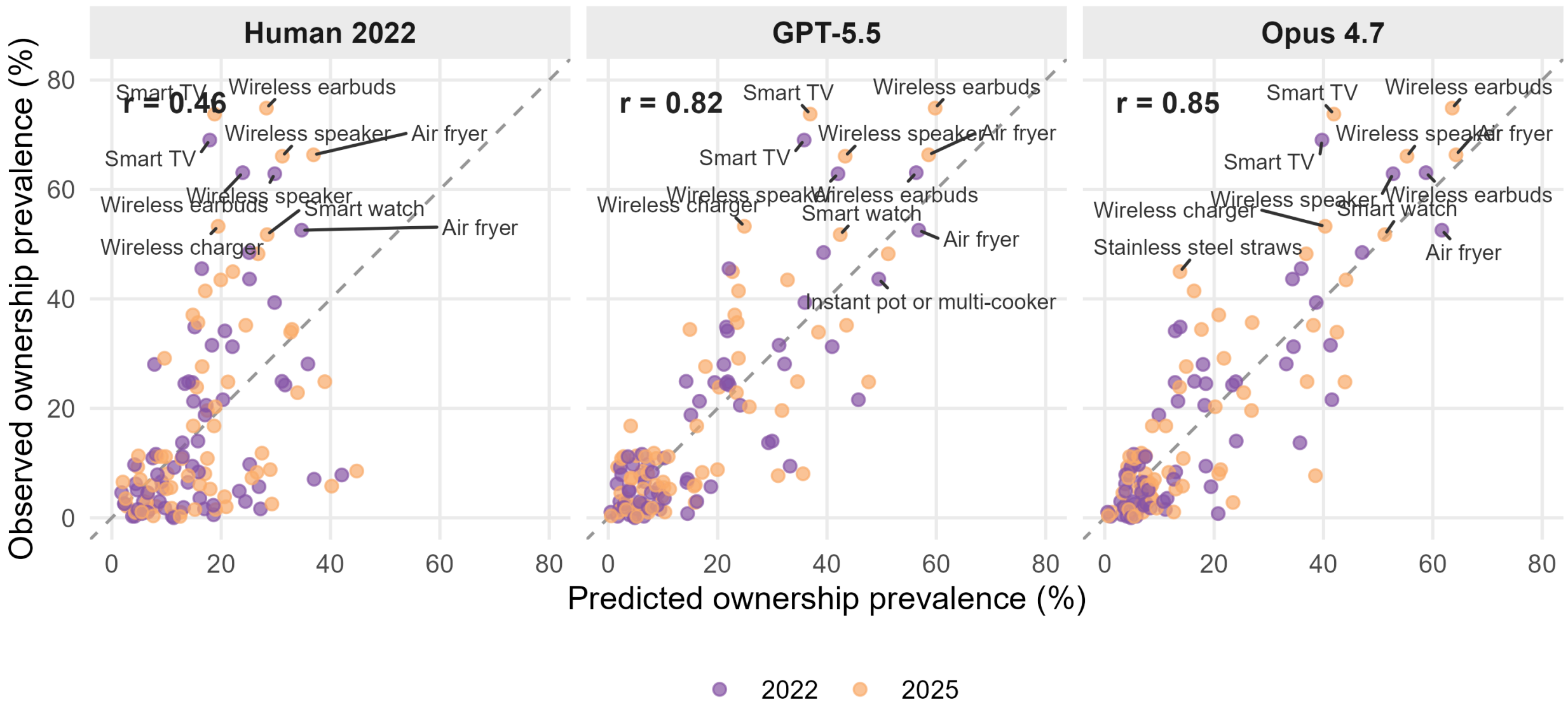

Figure 2: Observed vs predicted ownership prevalence (percentage points), one point per (technology, year) cell, under leave-one-technology-out. Panels: 2022 human raters (left), GPT-5.5 (centre), Opus 4.7 (right). Colour distinguishes the 2022 and 2025 surveys. Dashed line is the 45-degree identity. Pearson r per panel printed inside the panel.

## Which attributes carry the signal

Figure 3 shows the elastic-net coefficients on the four-attribute main cell, by rating source. Effort expectancy carries the largest positive coefficient on the logit scale for both model raters. For Opus 4.7 the remaining three attributes are smaller but not all zero, social influence ahead of performance expectancy and hedonic motivation. For GPT-5.5 all four attributes carry positive weight, with effort expectancy clearly on top. Effort expectancy and social influence are correlated across technologies, so the battery may divide a shared signal between them rather than mark two separable construct effects, and we interpret the coefficients as how the battery allocates weight rather than as the standalone contribution of each construct. The complete coefficient table, including the intercept, is in Table S4.

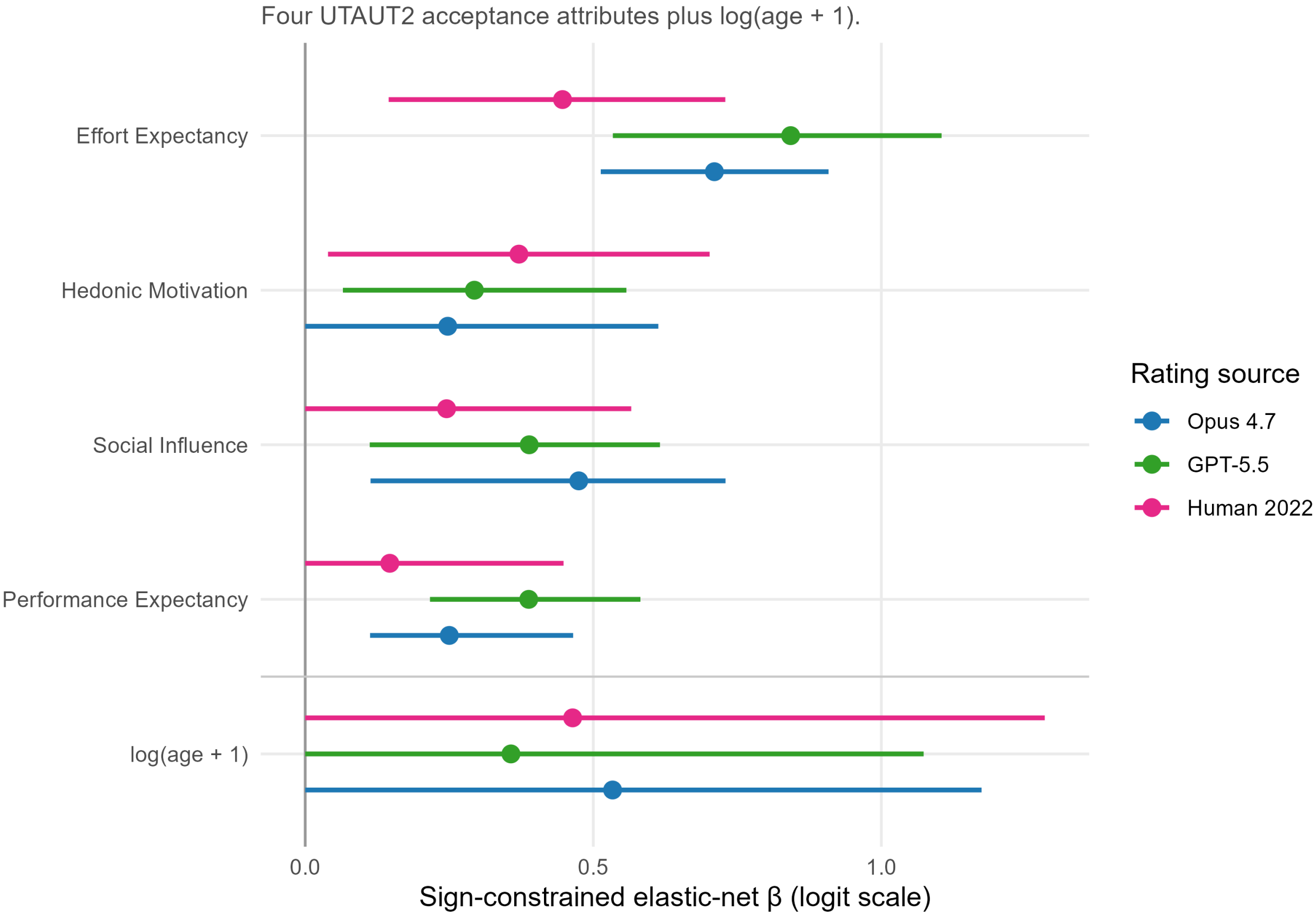

Figure 3: Sign-constrained elastic-net coefficients in the main cell, by rating source, with 95% bootstrap CIs over technology resamples (B = 1,000). The four UTAUT2 acceptance attributes (constrained ≥ 0, their UTAUT-theoretical direction) appear above the log(age + 1) slope (log_age constrained ≥ 0). Zero entries reflect either an active sign-constraint or elastic-net shrinkage. Coefficients are on the logit scale of ownership prevalence. Point estimates are fit on the full 65-technology sample at the penalty equal to the median, across leave-one-out folds, of the penalty chosen by inner cross-validation. The intercept is omitted here and reported in Table S4.

The human raters keep effort expectancy at the top of the ranking and divide the remaining weight more evenly across hedonic motivation, social influence, and performance expectancy, with no single attribute dominating. The human-rater cell has a higher MAE than the language-model cells in Figure 1. We examine the rater-source comparison, and the mechanisms that may account for it, in the Discussion.

## Three rating sources at the descriptive level

The MAE ordering Opus < GPT < Human in Figure 1 invites a descriptive comparison of the three rating sources, reported in Table S5 and Table S6. Three patterns matter for the Discussion. First, the model raters spread the technologies more widely on the 1–7 scale than the human raters do: averaged over the six attributes, the mean per-attribute standard deviation across the 65 technologies is 0.64 for Opus 4.7 and 0.96 for GPT-5.5, against 0.48 for the human raters, with the compression largest on facilitating conditions and hedonic motivation. Second, every attribute correlates more strongly with 2022 ownership for the model raters than for the human raters. Third, on social influence the two models agree with each other but barely with the human raters, where the cross-rater correlation is near zero, while the other attributes show moderate-to-high cross-rater agreement.

The human-rater disadvantage in Figure 1 might be an artifact of rating-sample size. A rarely owned technology rated by 9 respondents carries more measurement noise than one rated by 90, and small-sample averaging toward the grand mean would compress the across-technology variance in the human signal. To check this, we re-ran the main-model LOO on the 26 technologies recognized by at least 60% of the 2022 sample, each rated by at least forty respondents. The gap held: Opus 4.7 MAE 9.89 pp (95% CI [7.18, 13.01]) against human raters at 13.29 pp (95% CI [9.71, 16.97]), an absolute gap of 3.40 pp on the high-awareness subset versus 4.86 pp on the full set. Sample-size attenuation is therefore not what produces the cross-rater ordering, because the model advantage survives on exactly the technologies the human raters scored best. Absolute MAE rises for every rater on this subset, because removing the low-ownership tail, where all three cells are mechanically small, widens the range of outcomes.

Because the language-model raters scored each technology after the broad record of which products had become common was already public, a model could in principle let general knowledge of a technology's fate influence its attribute ratings. To test this, we compared an earlier and a later release of each model family on the same battery, asking whether the later release rates a known success higher and a known failure lower on the attributes associated with success. The knowledge-cutoff check looks for that pattern in three places and finds none. First, the baseline movement between releases: on the 65 training technologies, whose reception both releases already know, the six-attribute battery moved between releases by about 0.26 for Opus and 0.15 for GPT, roughly eleven and four times the movement expected from sampling alone, and 91% (Opus) and 55% (GPT) of technologies moved beyond their own sampling range. Because neither release could have learned about products on the market for a decade or more, this is recalibration between the model versions, the largest single source of change between the earlier and later releases. It is largest on social influence and on facilitating conditions. The size of that recalibration on facilitating conditions is one reason the attribute is kept in the supplement rather than the main model, so that the main result does not depend on the attribute that moves most between releases. Second, across the sixteen newer products, which were launched after the 2022 survey and are not part of the main analysis, where a release could genuinely have learned an outcome, the more a model's recall of a product changed across the cutoff, the less its attributes moved ($r = -0.65$ for Opus, no relationship for GPT), the opposite of what prior knowledge of outcomes would produce. Third, on the few products with a datable change in what the model knew about the outcome, the direction of movement is weak and mixed, with known successes and known failures shifting the same way rather than separating, the pattern expected under uniform late-version recalibration rather than ratings that follow the known outcome.

## Predicting the 2022-to-2025 change

The cross-sectional analysis collapses the two timepoints of data into one observation per technology-year. A more demanding test is whether the same attribute battery, combined with 2022 ownership as anchor, can predict where each technology stands in 2025 relative to where it was in 2022. Figure 4 reports leave-one-tech-out MAE on Δ-ownership in percentage points across 65 training technologies.

The zero-change baseline (predict 2025 ownership equals 2022 ownership) gives MAE 3.23 pp. Predicting the LOO mean change yields 3.02 pp. The anchor-only specification (logit-2022 + log age, no attributes), the baseline-plus-age cell in Figure 4, reaches 2.56 pp, numerically the best of the seven cells. Adding the four-attribute battery on top of the anchor returns Δ-MAE of 2.55 pp on the human-2022 ratings, 2.61 pp on GPT-5.5, and 2.59 pp on Opus 4.7. Across these 65 technologies the attribute battery does not improve on the anchor-only specification, and yields slightly higher MAE. The Pearson correlation between predicted and observed Δ on the Opus cell is 0.36 ($n = 65$, two-sided $p \approx 0.00$). A paired bootstrap (resample technologies, compute attribute-cell MAE minus zero-change MAE per resample, negative differences favouring the attribute model) puts the Opus advantage over predict-zero-change at −0.64 pp with 95% CI [−1.28, 0.01], a difference that crosses zero. All seven cells in Figure 4 fall inside a 0.68-pp band, well inside the bootstrap intervals on any single cell.

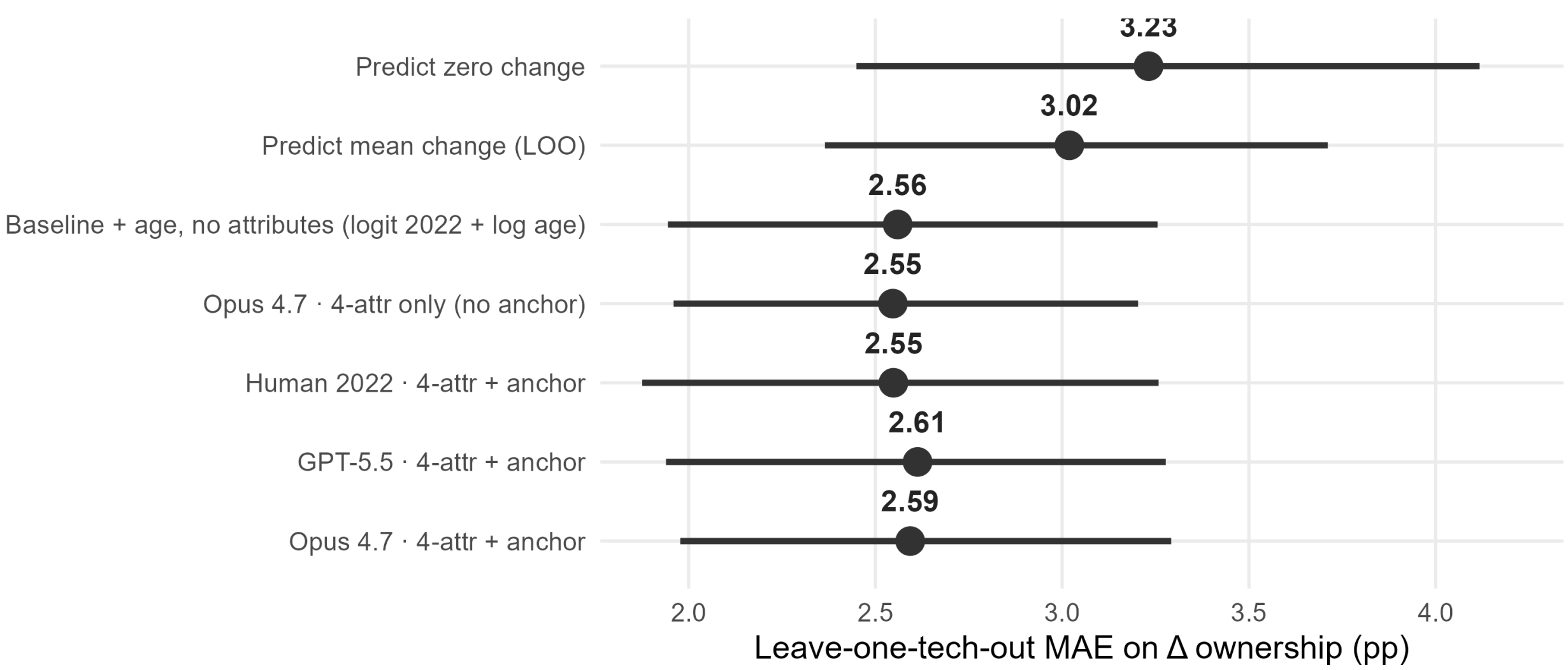

Figure 4: Leave-one-tech-out MAE on the 2022-to-2025 change in ownership prevalence (percentage points). The baseline-plus-age cell uses logit-2022 ownership and log age with no attribute ratings, and the three rated cells add the four-attribute battery for each rater. Bars are 95% bootstrap CIs over technologies.

# Illustrating deployment

To show how the model would be used on technologies outside the 2022 survey, we re-rated the 65 technologies together with eleven products launched in 2025 and 2026, refit the main specification on the re-rated technologies, and projected the eleven new products onto the same z-scaled coefficients. Table 1 reports the projected 2027 ownership share for each product, by rater. These eleven products range from premium home robots through wearables and privacy devices to a countertop cooking robot, chosen to span the expected range from genuine niche to mass-market candidate.

Table 1: Predicted ownership prevalence (%) for eleven illustrating-example consumer products in 2027, by language-model rater.

| Technology | Opus 4.7 pred. % | GPT-5.5 pred. % |
|---|---|---|
| Consumer ultrasonic chef's knife (Seattle Ultrasonics C-200) | 5.61 | 7.90 |
| Autonomous countertop cooking robot (Posha) | 4.22 | 4.01 |
| Bone-conduction music candy (Lollipop Star) | 2.77 | 4.43 |
| Microbe-based countertop composter (GEME Terra 2) | 2.16 | 3.00 |
| Pocket food-allergen detector (Allergen Alert) | 1.80 | 3.77 |
| In-ear EEG earbud (Naox Wave/Link) | 1.11 | 2.29 |
| Consumer humanoid home robot (1X Neo) | 1.04 | 0.44 |
| Soft-plastic home compactor (Clear Drop SPC) | 0.88 | 2.70 |
| Always-listening AI memory wristband (Bee Pioneer) | 0.58 | 1.38 |
| Ultrasonic anti-microphone audio-privacy shield (Deveillance Spectre I) | 0.47 | 0.40 |
| Consumer humanoid developer robot (Unitree R1) | 0.36 | 0.17 |

GPT-5.5 projects higher ownership than Opus 4.7 for most of the eleven new products, often several times higher, and the estimated coefficients account for the difference. GPT assigns positive weight to all four acceptance attributes, with effort expectancy the largest, whereas Opus relies most on effort expectancy. Both raters carry a positive log-age slope that pulls newly launched products down. The leave-one-out error in Figure 1 does not separate the two coefficient profiles, because each predicts its own technologies about equally well, and the profiles diverge only for products whose attribute or age values fall outside the range over which the coefficients were estimated. Which rating source comes closer to the eventual outcome cannot be determined within this sample, the limit of a single survey. With the coefficients fixed at their present values, a later study could evaluate these predictions against actual ownership once it is observed.

## Discussion

The present study tested whether the perceived attributes of a consumer technology, scored by a small set of raters, predict how widely it is owned, with no diffusion trajectory for the focal technology and no matched historical analogue. Two criteria framed the test. The first is out-of-sample prediction of cross-sectional ownership prevalence, measured against a covariate-only baseline that uses only years since a technology reached the market. The second is prediction of the 2022-to-2025 change in ownership, measured against a no-change baseline. The attribute battery predicts the level but not the three-year change, where it does not improve on the no-change baseline. That result is one the short window leaves underpowered rather than one the battery is shown to fail, and the contrast between the two outcomes marks where rated attributes carry signal and where the present design cannot yet settle the question. A third result concerns who supplies the ratings: the same four items predict ownership better when a language model supplies them than when human respondents do, with Opus ahead of GPT.

The cross-sectional result belongs to an established line of work that predicts adoption from product attributes rather than from an observed adoption curve. Lee and colleagues (2014) regressed Bass-curve parameters on expert-rated product and market attributes across a panel of products, Yamamura and colleagues (2022) combined expert domain knowledge with machine learning to forecast market share on automotive technologies, and the wider attribute tradition reaches back through UTAUT2 (Venkatesh, Thong, & Xu, 2012) to Rogers's perceived attributes of an innovation (Rogers, 1983). The present result agrees with that record: perceived attributes do carry adoption signal, and here that signal holds out of sample. The present design departs from this precedent in the source of its predictors and the breadth of its panel. The earlier cross-panel studies used expert-rated product attributes or expert-curated inputs within a single product category. The present mapping uses perception-based ratings across 65 technologies that span categories, and where UTAUT2 has served mostly to model one technology and individual acceptance, here the same constructs order population ownership across many. Because the mapping is fit across technologies rather than estimated from any single trajectory, it uses no matched analogue as input, the property that would let it reach a genuinely new technology for which no analogue exists.

The battery does not, however, predict the change over time. On the 2022-to-2025 change, the anchor-only specification of 2022 ownership plus log age is the best of the seven cells, and adding the four attributes leaves the error slightly higher rather than lower for every rater. A paired bootstrap on the difference from the no-change baseline crosses zero throughout. The window is part of the reason. Three years is short, most of the 65 technologies barely moved, and what variance a predictor could capture lies in the few technologies that changed substantially. We take this as a null on these data and over this short time window, not as evidence that rated attributes carry no dynamic signal at all. Separating the two would take a wider window or a set of technologies with larger shifts in ownership.

## Why the rater ordering goes Opus < GPT < Human

The descriptive comparison of the three rating sources narrows why the language-model ratings predict ownership better than the human raters do. The first candidate is design-driven attenuation of the human signal. Because only respondents who recognize a technology rate it, rarely owned technologies fall to small rating samples, and small-sample averages compress across-technology variance toward the grand mean. The human raters' attribute scores vary less across the 65 technologies than either model's, leaving a penalized regression less variation to exploit at every fold. The direct check is the high-awareness subset. Restricting the leave-one-out to the technologies recognized by most of the 2022 sample, each rated by at least forty respondents, leaves the rater gap essentially unchanged. Per-technology sample-size attenuation is therefore not the mechanism at work, because the language-model advantage holds on exactly the technologies where the human raters have their largest and least attenuated samples.

The second candidate is construct-validity drift, sharpest on social influence. The two models agree with each other on that item far more closely than either agrees with the human raters (Table S5). They appear to interpret it as a market-coverage question, that is, how widely a product is in use, while respondents interpret it as a social-network question, that is, how visibly people they know use it. The two interpretations converge on widely diffused technologies and diverge on niche ones. The split is weaker on facilitating conditions and absent on price value, which has a stable external referent. This matches the construct-validity gap in the language-model-as-rater literature (Bisbee et al., 2024; Goli & Singh, 2024): raters can agree numerically on a construct yet measure different latent quantities, with the gap widening where the construct rests on respondent-local social experience rather than an external anchor. The model cells predict contemporaneous ownership better here, but whether they measure the underlying attributes better is a separate question this design does not settle.

The third candidate is prior knowledge of outcomes. The language models rated each technology in 2026, when the broad record of which products had become common was public and in their training data. That record does not contain the present outcome, because the Prolific ownership shares here were never released, so the concern is not that a model recites the answer but that general knowledge of a technology's fate influences its attribute ratings. The knowledge-cutoff check is the direct test: the movement between the earlier and later releases reflects version recalibration, a larger change in what a model recalled about a product does not predict more movement in its attribute ratings, and facilitating conditions, the rated attribute whose availability for a new technology is most in question, shows no sign of following the known outcome. The check is suggestive rather than decisive, because it is a parallel rather than identical elicitation and cannot test the 65 main-text technologies directly, since both releases already knew which of them became common.

### Limitations

Two limitations bound the cross-sectional result. The first is structural: the leave-one-out scheme holds out a technology the raters had already encountered by 2026, so the result is a prediction for technologies already on the market, not a forecast of one not yet observed. Whether frozen coefficients can forecast a technology no rater has seen is the question the closing section returns to. The second is the outcome itself: ownership prevalence here is the share of this Prolific sample reporting ownership, so it is specific to this sample rather than a US-population share.

## Conclusions

The present study takes a first step toward attribute-only adoption forecasting rather than delivering it. It establishes a precondition for that goal: the mapping from a compact set of perceived attributes to population ownership is stable enough across heterogeneous technologies to predict the ownership of a held-out one. The underlying step is to interpret UTAUT2 across technologies rather than within a single one, because constructs validated for one person's acceptance of one technology also order population ownership across 65 products that span categories. The rater comparison adds that a frontier language model can supply those ratings, and that predictions built on them are at least as accurate as those from a fielded survey of human raters, better on these data. A knowledge-cutoff check finds no sign that prior knowledge of outcomes moved the ratings on the products where that knowledge could have changed, but it cannot test the long-known technologies behind the main result, so it narrows that explanation rather than excluding it. What it does not deliver is the forecast itself, and the three-year-change null marks where the attribute signal stops.

The present result is a prediction for a technology already on the market: it reproduces the adoption of a held-out technology whose place in the market is already settled. Whether the same mapping can forecast, reaching a technology before anyone owns it, is the harder question, and it is newly within reach, because a frontier model will score an unreleased product on the same battery in seconds. What no design has yet shown is whether attribute ratings, from human raters or a language model, carry the signal across that divide, from technologies the market has already judged to one it has not.

## Data and code availability

The aggregated survey data, the language-model ratings, and the Quarto engine that reproduces every figure, table, and number in this article are openly available on the Open Science Framework (https://osf.io/dr5ct).

# Supplementary material

## Supplementary Tables

**Table S1.** The 65 consumer technologies in the training sample, ordered from most to least owned in 2022. Type marks the survey instrument: AI-embedded products (25) or non-AI products (40). Introduced is the year consumer-grade products in the category first reached the US market, the basis for the log-age covariate. Owned and Aware are the aggregated Prolific ownership and awareness shares, each computed over every respondent who answered, the same unweighted basis as the main-text prevalences. Raters (n) is the 2022 per-technology rating-sample size. Because human raters scored only the technologies a respondent recognised, this count varies across products. Ownership runs from roughly two-thirds of respondents down to near zero, and the AI-embedded and non-AI products interleave across that range rather than separating into high- and low-penetration blocks.

| Technology | Type | Introduced | Owned 2022 (%) | Owned 2025 (%) | Aware 2022 (%) | Raters (n) |
|---|---|---|---|---|---|---|
| Smart TV | AI | 2010 | 69.0 | 73.8 | 91 | 37 |
| Wireless earbuds | non-AI | 2016 | 63.1 | 74.9 | 92 | 44 |
| Wireless speaker | non-AI | 2010 | 62.9 | 66.1 | 90 | 37 |
| Air fryer | non-AI | 2010 | 52.6 | 66.3 | 90 | 34 |
| Smart watch | AI | 2015 | 48.5 | 51.7 | 87 | 48 |
| Wireless charger | non-AI | 2015 | 45.5 | 53.3 | 85 | 41 |
| Instant pot or multi-cooker | non-AI | 2010 | 43.6 | 48.2 | 79 | 29 |
| Smart speaker | AI | 2014 | 39.3 | 33.9 | 76 | 47 |
| App-based live-TV | non-AI | 2015 | 34.9 | 41.5 | 71 | 41 |
| Stainless steel straws | non-AI | 2010 | 34.1 | 45.0 | 76 | 40 |
| Cordless vacuum cleaner | non-AI | 2010 | 31.5 | 43.5 | 79 | 41 |
| Smartphone grip (such as PopSockets) | non-AI | 2014 | 31.3 | 35.2 | 67 | 39 |

| Technology | Type | Intro-duced | Owned 2022 (%) | Owned 2025 (%) | Aware 2022 (%) | Raters (n) |
|---|---|---|---|---|---|---|
| Selfie stick | non-AI | 2014 | 28.1 | 24.9 | 89 | 36 |
| Weighted blanket | non-AI | 2017 | 28.0 | 29.1 | 77 | 42 |
| Mattress in a box | non-AI | 2014 | 24.9 | 35.7 | 65 | 89 |
| Touch-screen gloves | non-AI | 2009 | 24.9 | 34.4 | 60 | 38 |
| GPS tracker | non-AI | 2005 | 24.7 | 23.9 | 84 | 26 |
| Video door bell | AI | 2013 | 24.5 | 37.1 | 83 | 22 |
| No-contact fever thermometer | non-AI | 2010 | 24.3 | 22.9 | 64 | 75 |
| Robotic vacuum cleaner | AI | 2002 | 21.6 | 24.8 | 82 | 53 |
| Wi-Fi-controlled light bulb | non-AI | 2012 | 21.3 | 27.6 | 62 | 42 |
| Smart lighting | AI | 2012 | 20.6 | 20.3 | 70 | 43 |
| Smart plug | AI | 2012 | 18.8 | 16.8 | 47 | 38 |
| Bluetooth tracker (such as Tile) | non-AI | 2013 | 14.0 | 19.6 | 56 | 60 |
| Wearable health monitor | AI | 2009 | 13.7 | 7.7 | 54 | 37 |
| Smart home hub | AI | 2014 | 11.6 | 11.2 | 52 | 32 |
| VR headset with AI technology | AI | 2016 | 11.2 | 16.8 | 61 | 37 |
| Learning thermostat | AI | 2011 | 10.9 | 11.2 | 39 | 30 |
| Digital notepad | non-AI | 2017 | 9.8 | 11.8 | 54 | 45 |
| Mesh router | non-AI | 2016 | 9.7 | 11.3 | 30 | 40 |
| Clip-on selfie ring light | non-AI | 2015 | 9.4 | 8.0 | 44 | 38 |
| Smart TV antenna | non-AI | 2013 | 9.2 | 10.8 | 30 | 36 |
| Smart lock | AI | 2013 | 8.4 | 10.8 | 54 | 30 |

| Technology | Type | Intro-duced | Owned 2022 (%) | Owned 2025 (%) | Aware 2022 (%) | Raters (n) |
|---|---|---|---|---|---|---|
| Smart refrigerator | AI | 2016 | 7.9 | 11.2 | 66 | 37 |
| Digital pen (a.k.a. smart pen) | non-AI | 2008 | 7.8 | 8.5 | 46 | 36 |
| Waterproof e-reader | non-AI | 2014 | 7.0 | 5.8 | 36 | 33 |
| Intelligent alarm system | AI | 2013 | 6.6 | 8.0 | 37 | 29 |
| Self-heating mug | non-AI | 2017 | 6.5 | 6.0 | 43 | 47 |
| 3D printer | non-AI | 2009 | 6.2 | 9.3 | 84 | 85 |
| Automatic touchless hand-soap dispenser | non-AI | 2008 | 5.7 | 8.8 | 78 | 90 |
| Fingerprint padlock | non-AI | 2016 | 5.4 | 5.5 | 46 | 61 |
| Smart smoke detector | AI | 2013 | 5.1 | 7.0 | 38 | 31 |
| Smartphone projector | non-AI | 2012 | 4.9 | 7.3 | 40 | 43 |
| Smart oven | AI | 2015 | 4.6 | 5.9 | 40 | 26 |
| Hoverboard or self-balancing scooter | non-AI | 2014 | 4.6 | 6.5 | 76 | 84 |
| Smart pet cam | AI | 2014 | 3.6 | 5.2 | 42 | 21 |
| Nano-technology air purifier | non-AI | 2015 | 3.0 | 5.3 | 24 | 44 |
| Sunrise alarm clock | non-AI | 2006 | 3.0 | 8.3 | 42 | 40 |
| 3D-printing pen | non-AI | 2013 | 3.0 | 4.5 | 30 | 42 |
| Smart glasses | AI | 2013 | 2.6 | 3.5 | 39 | 27 |
| Drone with AI technology | AI | 2016 | 2.5 | 2.1 | 42 | 25 |
| Smart automatic pet feeder | AI | 2015 | 2.3 | 3.8 | 50 | 25 |
| Wearable directional-sound speaker | non-AI | 2017 | 1.9 | 1.5 | 19 | 28 |
| Smart nanny cam | AI | 2014 | 1.8 | 1.7 | 40 | 15 |

| Technology | Type | Intro-duced | Owned 2022 (%) | Owned 2025 (%) | Aware 2022 (%) | Raters (n) |
|---|---|---|---|---|---|---|
| Moldable glue (such as Sugru or Kintsuglue) | non-AI | 2009 | 1.6 | 2.5 | 15 | 31 |
| Automatic touchless toothpaste dispenser | non-AI | 2014 | 1.6 | 1.5 | 36 | 65 |
| AI toy robot | AI | 2016 | 1.5 | 2.1 | 35 | 25 |
| Robotic lawn mower | AI | 2015 | 1.5 | 1.0 | 33 | 20 |
| Robot nanny | AI | 2016 | 1.0 | 0.3 | 9 | 9 |
| Oura ring | AI | 2015 | 0.8 | 2.8 | 12 | 26 |
| Automatic handsfree toothbrush | non-AI | 2018 | 0.5 | 2.0 | 13 | 22 |
| Electronic posture trainer | non-AI | 2014 | 0.3 | 1.0 | 16 | 23 |
| Self-tying shoes (a.k.a. power laces) | non-AI | 2016 | 0.3 | 1.0 | 14 | 27 |
| Automatic pot stirrer | non-AI | 2014 | 0.3 | 0.8 | 22 | 36 |
| Window-cleaning robot | non-AI | 2010 | 0.0 | 0.3 | 14 | 20 |

**Table S2.** Two-level confirmatory factor analysis of the six-construct UTAUT2 attribute battery, estimated on the individual-level 2022 responses (2,540 complete batteries across 65 technologies) to check whether the survey items measure the intended UTAUT2 constructs. Individual raters are clustered by technology. The within-technology columns describe individual raters and the between-technology columns describe technologies. The battery measures the six constructs well within technologies, across individual raters, and is weaker across technologies, where 65 clusters are too few to evaluate against the usual between-level fit benchmark.

| Construct | Item | Est | SE | λ within [95% CI] | λ between | ICC | CR (w / b) | AVE (w / b) |
|---|---|---|---|---|---|---|---|---|
| Performance expectancy | PE1 | 0.62 | 0.02 | 0.54 [0.51, 0.57] | 0.58 | 0.17 | 0.83 / 0.92 | 0.64 / 0.81 |
| | PE2 | 1.21 | 0.02 | 0.87 [0.86, 0.89] | 1.00 | 0.17 | | |
| | PE3 | 1.22 | 0.02 | 0.87 [0.85, 0.88] | 0.99 | 0.18 | | |
| Hedonic motivation | HM1 | 1.10 | 0.02 | 0.88 [0.86, 0.89] | 1.00 | 0.11 | 0.84 / 0.94 | 0.65 / 0.84 |
| | HM2 | 0.85 | 0.02 | 0.75 [0.72, 0.77] | 0.87 | 0.07 | | |
| | HM3 | 1.04 | 0.02 | 0.78 [0.76, 0.79] | 0.88 | 0.16 | | |
| Effort expectancy | EE1 | 0.95 | 0.02 | 0.84 [0.83, 0.86] | 1.00 | 0.22 | 0.87 / 0.99 | 0.68 / 0.98 |
| | EE2 | 0.88 | 0.02 | 0.80 [0.78, 0.81] | 0.97 | 0.17 | | |
| | EE3 | 0.96 | 0.02 | 0.84 [0.83, 0.86] | 1.00 | 0.23 | | |
| Social influence | SI1 | 1.07 | 0.02 | 0.77 [0.75, 0.79] | 1.00 | 0.05 | 0.83 / 0.99 | 0.61 / 0.96 |
| | SI2 | 0.92 | 0.02 | 0.75 [0.73, 0.77] | 1.00 | 0.08 | | |
| | SI3 | 1.12 | 0.02 | 0.82 [0.80, 0.84] | 0.95 | 0.05 | | |
| Facilitating conditions | FC1 | 0.83 | 0.02 | 0.73 [0.70, 0.76] | 0.61 | 0.18 | 0.54 / 0.57 | 0.29 / 0.32 |
| | FC2 | 0.52 | 0.03 | 0.41 [0.38, 0.45] | 0.38 | 0.14 | | |
| | FC3 | 0.57 | 0.03 | 0.46 [0.42, 0.50] | 1.00 | 0.03 | | |
| Price value | PV1 | 1.11 | 0.02 | 0.85 [0.83, 0.86] | 1.00 | 0.12 | 0.89 / 1.00 | 0.74 / 1.00 |
| | PV2 | 1.07 | 0.02 | 0.84 [0.83, 0.86] | 1.00 | 0.11 | | |
| | PV3 | 1.18 | 0.02 | 0.88 [0.87, 0.89] | 1.00 | 0.12 | | |

*Notes.* Est and SE are the unstandardized factor loading and its standard error, common to both levels because loadings are held equal across levels. λ within and λ between are the standardized loadings at the within-technology and between-technology levels. CR and AVE are the composite reliability and average variance extracted per level (within / between). Composite reliability follows the multilevel framework of Geldhof et al. (2014) and average variance extracted follows Fornell and Larcker (1981). ICC is the intraclass correlation, the share of an item's variance that lies between technologies. The model was estimated by maximum likelihood, with within-level latent variances fixed to one and between-level residual variances fixed to zero for ten of the eighteen items, leaving free the eight items that retained non-trivial between-technology residual variance. All factor loadings are significant at $p < .001$. The model fits the within-technology level well (CFI = 0.94, RMSEA = 0.05, within-level SRMR = 0.06). The between-technology SRMR is higher (0.21), as expected with only 65 technology clusters, a sample size at which this index is biased upward and cannot be read against the conventional 0.08 cutoff (Asparouhov & Muthén, 2018). Convergent validity is adequate at the within level for five of the six constructs, with composite reliabilities from 0.83 to 0.89 and average variance extracted from 0.61 to 0.74. Facilitating conditions is the weak construct (CR = 0.54, AVE = 0.29), and Cronbach's alpha shows the same pattern (0.81 to 0.91 for the other five constructs, 0.55 for facilitating conditions). Discriminant validity holds among the well-measured constructs, with the exception of facilitating conditions, which is weakly defined and not separable from effort expectancy at the within level (latent correlation 1.00). Across technologies the effort-expectancy, social-influence, facilitating-conditions, and price-value constructs converge toward a single dimension (between-level latent correlations from 0.72 to 0.97), with hedonic motivation standing apart, so we do not interpret the individual attribute coefficients in the main model as separable contributions of distinct properties. A strict specification that fixes all between-level residual variances to zero fits the present data less well (CFI = 0.90, between-level SRMR = 0.25). PE performance expectancy, HM hedonic motivation, EE effort expectancy, SI social influence, FC facilitating conditions, PV price value.

**Table S3.** Attribute item wording in the 2022 instrument, by construct. Each construct was a three-item composite rated 1 (*Strongly disagree*) to 7 (*Strongly agree*), under the third-person framing "how people in general perceive [technology]", with the product name piped into each item in place of [technology]. Owners and non-owners saw the same items apart from minor grammatical variants (for example, "helps" in the AI instrument and "can help" in the non-AI instrument).

| Construct | Item wording |
|---|---|
| Performance expectancy | [technology] can be useful in people's daily life |
| | [technology] helps people accomplish things more quickly |
| | [technology] helps people carry out tasks more quickly |
| Hedonic motivation | For people in general, [technology] is fun to use |

| Construct | Item wording |
|---|---|
| | People enjoy using [technology] |
| | [technology] is entertaining to use, for people in general |
| Effort expectancy | People in general find [technology] easy to use |
| | For people in general, interacting with [technology] is clear and understandable |
| | Learning how to use [technology] is easy, for people in general |
| Social influence | [technology] is a product that people use if others around them prefer them to |
| | [technology] is a product that people use if others recommend them to |
| | [technology] is a product that people use if others think that they should |
| Facilitating conditions | People in general have the knowledge necessary to use [technology] |
| | [technology] is compatible with other technologies that people tend to use |
| | When someone has trouble using [technology], they can get help |
| Price value | People in general perceive [technology] as reasonably priced |
| | People in general perceive [technology] as good value for money |
| | At the current price, people think that [technology] is good value for money |

**Table S4.** All coefficients in the main four-attribute cell (PE, HM, EE, SI, with no facilitating conditions and no price value), by rating source. Point estimates are fit on the full 65-technology sample at the median, across leave-one-out folds, of that inner-cross-validation penalty. Bracketed values are 95% bootstrap confidence intervals over technology resamples (B = 1,000). Coefficients are on the logit scale of ownership prevalence. Zero entries reflect either an active sign-constraint (acceptance attributes constrained ≥ 0) or elastic-net shrinkage. The 'Block' column separates the four UTAUT2 attribute coefficients from the structural controls (intercept and log(age + 1)) that enter the same fit. Figure 3 plots all of these except the intercept, which Table S4 adds.

| Block | Term | Opus 4.7 | GPT-5.5 | Human 2022 |
|---|---|---|---|---|
| Attribute coefficient | Performance Expectancy | 0.250 [0.113, 0.465] | 0.388 [0.217, 0.582] | 0.147 [0.000, 0.448] |
| Attribute coefficient | Hedonic Motivation | 0.247 [0.000, 0.613] | 0.294 [0.065, 0.557] | 0.371 [0.040, 0.702] |

| Block | Term | Opus 4.7 | GPT-5.5 | Human 2022 |
|---|---|---|---|---|
| Attribute coefficient | Effort Expectancy | 0.710 [0.513, 0.908] | 0.843 [0.534, 1.105] | 0.447 [0.145, 0.729] |
| Attribute coefficient | Social Influence | 0.475 [0.113, 0.730] | 0.389 [0.112, 0.616] | 0.245 [0.000, 0.566] |
| Structural control | Intercept | −3.251 [−4.920, −1.997] | −2.876 [−4.708, −1.873] | −2.925 [−4.963, −1.674] |
| Structural control | log(age + 1) | 0.534 [0.000, 1.174] | 0.357 [0.000, 1.073] | 0.464 [0.000, 1.284] |

*Notes.* Each cell reports the point estimate followed by the 95% bootstrap confidence interval in brackets, both on the logit scale. The bootstrap resamples the 65 training technologies with replacement (B = 1,000), refits the elastic-net at the same penalty, and takes the 2.5% and 97.5% quantiles of each coefficient across resamples. The three raters reach similar leave-one-out MAE on the 65-technology sample (Figure 1) but allocate that fit across the attribute and structural terms differently. The trade-off is observationally indistinguishable on the 130 training cells and surfaces only when the model is projected onto products whose attribute or age profile lies outside the range over which the coefficients were estimated (see the Table 1 example in the main text).

**Table S5.** Cross-rater Pearson correlations on individual UTAUT attributes, computed over the 65 training technologies (2022 survey), on all six rated attributes (the four main acceptance constructs, facilitating conditions, and price value). Each entry is the Pearson correlation between a pair of rating sources. Agreement is strongest on price value, which has a relatively stable external referent across raters, and weakest on social influence, where the language-model and human raters are not measuring the same latent quantity.

| Attribute | Opus–GPT r | Opus–Human r | GPT–Human r |
|---|---:|---:|---:|
| Performance Expectancy | 0.88 | 0.66 | 0.63 |
| Hedonic Motivation | 0.92 | 0.71 | 0.76 |
| Effort Expectancy | 0.95 | 0.85 | 0.81 |
| Social Influence | 0.85 | 0.03 | 0.11 |
| Facilitating Conditions | 0.88 | 0.66 | 0.68 |
| Price Value | 0.96 | 0.73 | 0.72 |

*Notes.* A separate re-elicitation diagnostic speaks to the reliability of the language-model ratings. Re-running the full battery in a later, larger batch reproduced the original 65-technology values closely, with batch-to-batch Pearson correlations of at least 1.00 on every attribute except social influence (1.00 for Opus 4.7, 1.00 for GPT-5.5) and median absolute deviations below half a point

on the 1–7 scale throughout. Social influence is again the least stable construct, in line with the cross-rater pattern.

**Table S6.** Per-rater descriptive comparison on the 65 training technologies (2022 survey). SD is the standard deviation of each attribute's per-technology means, computed across technologies on the 1–7 attribute scale. r(own) is the Pearson correlation between the per-technology attribute mean and 2022 Prolific ownership prevalence. These are the per-rater descriptive numbers used in the main-text descriptive comparison and Discussion.

| Attribute | Opus 4.7 SD | Opus 4.7 r(own) | GPT-5.5 SD | GPT-5.5 r(own) | Human 2022 SD | Human 2022 r(own) |
|---|---|---|---|---|---|---|
| Performance Expectancy | 0.68 | 0.46 | 0.98 | 0.36 | 0.57 | 0.32 |
| Hedonic Motivation | 0.69 | 0.42 | 1.03 | 0.32 | 0.48 | 0.33 |
| Effort Expectancy | 0.66 | 0.54 | 0.99 | 0.58 | 0.59 | 0.42 |
| Social Influence | 0.41 | 0.67 | 0.72 | 0.58 | 0.38 | 0.38 |
| Facilitating Conditions | 0.66 | 0.78 | 0.87 | 0.63 | 0.33 | 0.50 |
| Price Value | 0.73 | 0.72 | 1.16 | 0.71 | 0.51 | 0.59 |

## Supplementary Figures

**Figure S1.** Leave-one-technology-out mean absolute error (MAE) on Prolific 2022+2025 ownership prevalence, per rating source, for three specifications of the attribute battery. Baseline is the four UTAUT2 acceptance attributes in the main model (performance expectancy, hedonic motivation, effort expectancy, and social influence). The two add-on cells extend the baseline with facilitating conditions or with price value. The Opus 4.7 baseline cell, the main model reported in the Results (Figure 1), is marked with an asterisk. Bars are 95% bootstrap intervals over technology resamples.

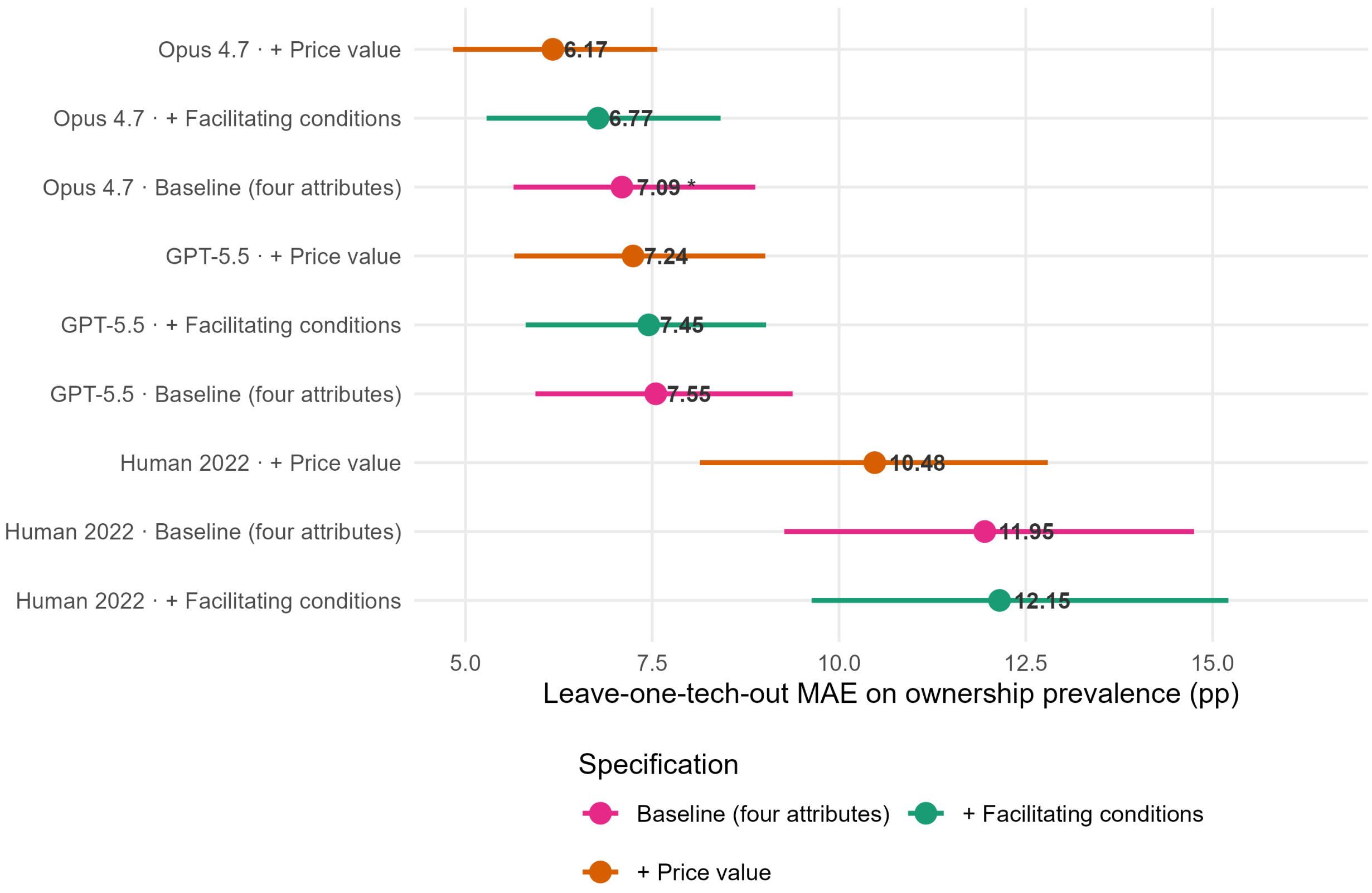

*Notes.* Facilitating conditions is the attribute most plausibly downstream of adoption (see Discussion), so the baseline drops it from the main model. Effort expectancy and the other acceptance attributes absorb most of the facilitating-conditions signal, and adding it back as the second cell moves held-out error only within the bootstrap interval, so the cross-sectional result does not depend on the one attribute most exposed to the endogeneity concern. PE performance expectancy, HM hedonic motivation, EE effort expectancy, SI social influence, FC facilitating conditions, PV price value.